\documentclass[graybox]{svmult}

\usepackage{mathptmx}       % selects Times Roman as basic font
\usepackage{helvet}         % selects Helvetica as sans-serif font
\usepackage{courier}        % selects Courier as typewriter font
\usepackage{type1cm}        % activate if the above 3 fonts are
\usepackage{makeidx}         % allows index generation
\usepackage{graphicx}        % standard LaTeX graphics tool
\usepackage{multicol}        % used for the two-column index
\usepackage[bottom]{footmisc}% places footnotes at page bottom

\usepackage{hyperref}
\usepackage{url}

\makeindex             % used for the subject index
\begin{document}

%\title*{Detecting the Invisible: Intelligent Structural Monitoring with Shearography}
\title*{Automated Annotation of Shearographic Measurements Enabling Weakly Supervised Defect Detection
\thanks{This preprint has not undergone peer review or any post-submission improvements or corrections. 
    		The Version of Record of this contribution will be published in:
            Complex, Intelligent and Software Systems. Lecture Notes in Data Engineering and Communication Technologies (Series ID: 15362). Springer Nature Switzerland AG, in press.}}

\titlerunning{Zero-Shot Annotation for Shearography} 

\author{Jessica Plassmann and Nicolas Schuler and Michael Schuth and Georg von Freymann}
\authorrunning{J. Plassmann et al.} %for an abbreviated version of

\institute{Jessica Plassmann
\at Department of Engineering and technology center OGKB, Trier University of Applied Sciences, Germany \email{plassmaj@hochschule-trier.de} 
\at Department of Physics and research center OPTIMAS, RPTU University Kaiserslautern-Landau, Germany
\and Nicolas Schuler 
\at Department of Computer Science and Laboratories for Cognitive Systems and Robotics, Trier University of Applied Sciences, Germany 
\email{schulern@hochschule-trier.de}
\at Department of Engineering, Faculty of Science, Technology and Medicine, University of Luxembourg, Luxembourg
% \and Valentin Bastgen 
% \at Department of Engineering and technology center OGKB, Trier University of Applied Sciences, Germany \email{plassmaj@hochschule-trier.de} 
% \at Department of Physics, RPTU University Kaiserslautern-Landau, Germany
\and Michael Schuth 
%\at Name, Address of Institute \email{name@email.address}
\at Department of Engineering and technology center OGKB, Trier University of Applied Sciences, Germany \email{schuth@hochschule-trier.de} 
\and Georg von Freymann 
\at Department of Physics and research center OPTIMAS, RPTU University Kaiserslautern-Landau, Germany \email{georg.freymann@rptu.de}
\at Fraunhofer Institute for Industrial Mathematics ITWM, Kaiserslautern, Germany
}
\maketitle
\abstract*{Shearography is an interferometric technique sensitive to surface displacement gradients, providing high sensitivity for detecting subsurface defects in safety-critical components. A key limitation to industrial adoption is the lack of high-quality annotated datasets, since manual labeling remains labor-intensive, subjective, and difficult to standardize.
We present an automated labeling pipeline that generates candidate defect bounding boxes with Grounded DINO, refines them using SAM masks, and exports YOLO-format labels for downstream detector training. Quantitative evaluation shows the generated boxes are suitable for weakly supervised learning, while high-resolution masks provide qualitative visualization. This approach reduces manual effort and supports scalable dataset creation for robust industrial defect detection.}

\abstract{Shearography is an interferometric technique sensitive to surface displacement gradients, providing high sensitivity for detecting subsurface defects in safety-critical components. A key limitation to industrial adoption is the lack of high-quality annotated datasets, since manual labeling remains labor-intensive, subjective, and difficult to standardize.
We present an automated labeling pipeline that generates candidate defect bounding boxes with Grounded DINO, refines them using SAM masks, and exports YOLO-format labels for downstream detector training. Quantitative evaluation shows the generated boxes are suitable for weakly supervised learning, while high-resolution masks provide qualitative visualization. This approach reduces manual effort and supports scalable dataset creation for robust industrial defect detection.}

\section{Introduction}\label{sec:Introduction}
    Shearography, or Electronic Speckle-Pattern Shearing Interferometry (ESPSI), is a non-contact, nondestructive testing (NDT) technique capable of measuring surface displacement gradients. It is particularly suited for detecting subsurface defects such as delaminations, voids, and microcracks in composites and other complex materials~\cite{Tao:22a, Petry:21}. Consequently, shearography is a promising tool for industrial quality assurance, especially in safety-critical areas.
    
    Despite the advantages of shearography, one of the factors limiting its industrial adoption is the lack of sufficiently automated data analysis. Modern deep learning techniques have the potential to significantly advance automation in defect detection~\cite{Bai2024_MLDMaterialDefectDetection}. However, their application to shearography presents specific challenges. Supervised approaches require large, high-quality annotated datasets for defect classification and localization~\cite{Hussain:23,YOLO:16}. Creating such datasets is labor-intensive, time-consuming, and difficult to standardize, while manual labeling is prone to errors and subjective inconsistencies~\cite{Schmarje:21}. Synthetic datasets have been proposed to alleviate these limitations and have shown strong performance in various computer vision domains, particularly when combined with fine-tuning on real data~\cite{DeTone2018_SuperPoint,Wen2025_FoundationStereo}. However, in shearography, previous studies have shown that datasets comprising only a few hundred real measurements achieved superior defect localization compared to mixed real–synthetic datasets~\cite{Li:22}. This suggests that, while synthetic pretraining can be highly effective in general, generating physically realistic data for shearographic patterns remains particularly challenging.
    
    Unsupervised methods aim to reduce reliance on labeled data by identifying deviations from the mean of the training data, which in this context corresponds to a defect-free baseline. While these approaches eliminate the need for manual labeling, they are generally limited to binary classification or exhibit reduced performance when distinguishing multiple defect types or subtle variations in component behavior~\cite{Cui2023_UnsupervisedAnomalyDetection}. As a result, complex tasks such as multi-class defect classification or precise defect localization remain challenging without supervised training on specific datasets.
    
    The generation of such datasets continues to be a primary challenge for the field. To address this, we propose an automated annotation workflow that has been developed and validated for shearography, but is conceptually applicable to other imaging-based inspection domains. Grounded DINO~\cite{Liu2023_GroundingDINO} generates bounding boxes from textual prompts, which are subsequently refined into high-resolution segmentation masks using the Segment Anything Model (SAM)~\cite{Kirillov2023_SegmentAnything}. This pipeline enables reproducible and scalable dataset creation, reducing manual labeling effort while supporting weakly supervised training for downstream defect detection models.
    
    By facilitating automated annotation of shearography data, this approach provides a pathway toward efficient, reproducible, and high-resolution defect inspection. It demonstrates potential for deployment in industrial quality assurance, particularly in high-volume production environments such as the automotive industry, where reproducibility and efficiency are critical. 
    A preprint version of this work is available on \href{https://arxiv.org/abs/2512.06171}{arXiv:2512.06171}.

\section{Foundations and Related Work}\label{sec:Foundations}
    To contextualize the proposed annotation workflow, this section reviews the key concepts of shearography and zero-shot detection relevant for automated defect localization.
    
\subsection{Electronic Speckle Pattern Shearing Interferometry}
    In shearography, an optically rough surface is illuminated with coherent light, producing a speckle pattern that contains phase information. In the interferometric setup, the backscattered wavefront is laterally displaced (sheared), such that interference occurs between neighboring points on the specimen surface. This enables measurement of displacement gradients between different loading states, which are recorded in differential phase maps, or shearograms.~\cite{Nakadate:80}
    
    Quantitative evaluation requires phase extraction, which is typically realized using temporal or spatial phase-shifting techniques. Temporal phase shifting is achieved by varying the optical path length of the unsheared beam, for example with a piezo actuator. This method provides high accuracy but requires sequential image acquisition, which complicates the analysis of dynamic excitations such as thermal loading, where for example Lock-in methods must be applied~\cite{Menner2011_LockinInterferometry}. Spatial phase shifting, by contrast, introduces a carrier frequency via an aperture. Although this reduces light efficiency and precision, it enables single-shot acquisition and is therefore particularly advantageous for measurements under dynamic loading and for integration into serial industrial inspection.~\cite{Petry:21}
    
    Several interferometric configurations have been established, which mainly differ in the optical elements used to introduce the lateral displacement, such as for example biprisms, Wollaston prisms, or tilted mirrors, and in the components required to realize the desired phase shifting method~\cite{Schuth:95}. Among these, the Michelson design~\cite{Xie:13,Zhang2020_ModifiedMichelsonInterferometer} is compact and robust, making it well suited for industrial applications~\cite{Sirohi2022_ShearographyReview}. The Mach–Zehnder design~\cite{Cai:12} represents a further development particularly advantageous for spatial phase shifting shearography, since the relative arrangement of the aperture and the shear mirror decouples the aperture geometry from the applied shear~\cite{Petry:20}. This feature allows the shear to be adjusted more freely, which is especially beneficial in research and laboratory environments. The integration of spatial light modulators in modern setups further expands the range of experimental configurations~\cite{Wang:23}.
    
    While these configurations, excitation methods, and phase-shifting strategies vary, properly aligned systems produce comparable shearograms. The present work focuses on spatial phase-shifting shearography under thermal excitation, but the proposed annotation approach is broadly applicable across setups.
    
\subsection{Zero-Shot Detection}
    Recent advances in foundation models have introduced new paradigms for object detection and segmentation without task-specific training, commonly referred to as zero-shot detection. These approaches rely on large-scale transformer architectures trained in extensive image-text pairs, allowing the extraction of object masks or bounding boxes based solely on textual or visual prompts~\cite{Jain2023_OneFormer, Zou2023_SegmentEverything, Zhang2024_PersonalizeSAM}.
    
    A milestone in this development is SAM~\cite{Kirillov2023_SegmentAnything}, trained on more than one billion segmentation masks across eleven million images. SAM consists of an image encoder that generates embeddings, and a lightweight prompt encoder and mask decoder that allow efficient querying via prompts such as points, bounding boxes, or text. While the publicly available implementation currently exclusively supports point and box prompts, text-based queries can be integrated by coupling SAM with external zero-shot detectors such as Grounded DINO (OWL-ViT)~\cite{Minderer2022_SimpleOpenVocabularyDetection}, which generate bounding boxes from natural language input.
    
    The release of SAM has spurred a variety of applications, ranging from image editing~\cite{Yu2023_InpaintAnything} and object detection~\cite{Ye2024_SAMInstanceSegmentation} to anomaly detection in medical imaging~\cite{Zhang2024_SAMMedical}. For an overview of these developments, see~\cite{Zhang2023_SAMSurvey}. More recently, SAM 2~\cite{Ravi2025_SAM2} extended this framework to video processing by introducing mechanisms for propagating and updating segmentation masks across consecutive frames, enabling temporal tracking without repeated prompting.
    
    SAM 3~\cite{Carion2025_SAM3} expands the model family by introducing open-vocabulary detection and exemplar-based prompting, enabling the model to identify and segment all instances of a specified category in both images and videos. In addition to point, box and mask inputs, SAM 3 supports short text descriptions and visual exemplars, which are processed within a unified prompting framework. This extends the visually prompted segmentation and video tracking capabilities of earlier versions toward category-level retrieval based on language or example regions.
    
    In the context of shearography, zero-shot models offer a pathway to automated annotation and defect localization without the need for large, manually labeled datasets. By combining text-based region proposals with high-resolution segmentation, they support efficient and reproducible dataset creation, addressing a major bottleneck in industrial defect detection.

\section{Methodology}\label{sec:Methodology}
    The following sections describe the architecture of the proposed zero-shot annotation pipeline, its implementation, the dataset preparation and training setup, and the metrics used for quantitative evaluation.
    
    \subsection{Pipeline Architecture}
    \begin{figure*}[t]
        \centering
        \includegraphics[width=\linewidth]{Figures/zeroshotpipeline_2.pdf}
        \caption{Overview of the zero-shot annotation pipeline. Bounding boxes are generated by a pre-trained zero-shot detector (a), refined into segmentation masks by a zero-shot segmentor (b), and subsequently used for supervised training of defect detection models (c).}
        \label{fig:zeroshot_pipeline}
    \end{figure*}

    The proposed zero-shot annotation pipeline, illustrated in Fig.~\ref{fig:zeroshot_pipeline}, enables automated generation of high-quality defect labels without domain-specific training. This approach addresses the bottleneck associated with manual annotation of shearography measurements, allowing subsequent training of conventional defect detection models for real-time applications.
    
    The pipeline consists of three sequential stages:
    (a) Zero-shot detection: A pre-trained zero-shot model receives a prompt describing the defect of interest and outputs bounding boxes with associated confidence scores for all detected anomalies. This stage requires no domain-specific training and operates fully independently of the specific shearography setup.
    (b) Segmentation refinement: The bounding boxes are provided as input to the Segment Anything Model, which generates high-resolution segmentation masks corresponding to the detected regions. This step enhances the spatial precision of the annotations while remaining fully automated.
    (c) Supervised model training: The automatically generated masks serve as training labels for downstream defect detection models. Although this stage involves conventional supervised learning, the overall approach is considered unsupervised because the extraction of labels from raw data is fully automated.
    
\subsection{Implementation}
    The implementation of the zero-shot pipeline depends on two key components: the zero-shot detector used to generate bounding boxes and the SAM implementation employed for producing segmentation masks. In this work, Grounded DINO~\cite{Liu2023_GroundingDINO} with DINO~1.0~\cite{Zhang2023_DINO} was selected as the zero-shot detector, while SAM~1~\cite{Kirillov2023_SegmentAnything} served as the segmentation model. Both pre-trained networks were accessed through the Hugging Face Transformers library~\cite{Wolf2019_Transformers}, and data handling was implemented using PyTorch~\cite{PyTorch}.  
    
    % The pipeline applies knowledge distillation to transfer the capabilities of large zero-shot models into compact, task-specific detectors. Using an efficient DINO–SAM configuration for annotation generation, the resulting YOLO-based models support real-time inference on resource-limited hardware, enabling practical deployment in industrial settings. By capturing the behavior of prompt-dependent zero-shot models in explicitly trained detectors, reliance on expert-crafted textual prompts is eliminated. The framework is flexible and can be adapted to alternative model configurations if needed.
    
    Bounding boxes and segmentation masks are exported in YOLO format. Bounding boxes can originate either directly from Grounded DINO or indirectly from the segmentation masks produced by SAM, which naturally introduces subtle differences between the two types of annotations. Consequently, a single text prompt may yield three distinct label sets: bounding boxes from DINO, segmentation masks from SAM, and bounding boxes derived from those masks. For the downstream supervised stage, YOLOv8~\cite{YOLOv8} was adopted as the defect detection model, given its proven efficiency in real-time applications and wide adoption in industrial inspection tasks~\cite{Yu2025_EnhancedYOLOv8}.
    
    To improve robustness, heuristic filters are applied to the raw outputs. Only bounding boxes smaller than 20~\% of the image size are retained, and segmentation masks must cover at least 500~pixels to be considered valid. In addition, non-maximum suppression~\cite{Neubeck2006_NMS} is applied to reduce overlapping detections. These heuristics are parameterized and can be adapted to different datasets and application scenarios, ensuring stable annotation quality across varying measurement conditions.
    
    All experiments were performed on a Windows 11 workstation equipped with an Intel Core i7-13700 CPU, 128~GB RAM, and an NVIDIA RTX 4000 GPU with 16~GB VRAM. Models were implemented in Python 3.11 with CUDA~\cite{cuda} acceleration. 

\subsection{Dataset Preparation and Training Setup}
    The dataset used in this study is based on the publicly available SADD dataset~\cite{plassmann2025sadd}. Specifically, shearography measurements from the faulty category were selected, together with the corresponding expert annotations provided in the dataset. In total, 4310 measurements with bounding box annotations of the detected defects were used. Each sample includes metadata describing the measurement process and the annotated defects.  
    
    The measurements are organized into groups of four images, each representing the same specimen region under different temporal conditions. In total, 29 annotation classes were defined, including 25 distinct defect types (e.g. delaminations, voids, microcracks) and four auxiliary classes describing artifacts such as edge effects or unintentional perturbations. For the present work, the 25 defect classes were consolidated into a single generic defect class, while the auxiliary classes were excluded from further analysis.  
    
    The dataset was split into training, validation, and test partitions of 3523, 391, and 396 images, respectively. To avoid data leakage, all four images of each group were assigned to the same partition. Furthermore, each subset was constructed to ensure a minimum of 20 occurrences of every defect class across the training, validation, and test data. In addition, 254 images without defects were added to the test set, providing a representative sample of defect-free cases for binary classification evaluation. 
    
    For evaluation of the zero-shot annotation pipeline, two textual prompts, “Cells” and “Two Circles”, were chosen based on prior studies identifying them as the most effective descriptors for the defects of interest. As the underlying zero-shot models are generic and not trained on shearographic data, task-specific terminology is not recognized. Thus, prompt engineering is required to identify textual formulations that effectively map to the visual representations learned by the model. For each prompt, three types of annotations were generated and employed to train YOLOv8 models for defect detection. The first type consists of bounding boxes directly predicted by Grounded DINO, the second type comprises segmentation masks produced by SAM, and the third type contains bounding boxes derived from the corresponding SAM segmentation masks. In addition, manually labeled bounding boxes were included as ground truth references, referred to as YOLOHand.
    
    The code and all associated results are publicly available in the corresponding repository~\cite{Git:Data}.
    
\subsection{Evaluation Metrics}
The evaluation of the proposed method focuses on two complementary aspects: binary classification between defect-free and defective samples, and localization accuracy. All metrics were computed using the TorchMetrics library~\cite{Detlefsen:22}.

Binary classification performance was assessed using the Receiver Operating Characteristic (ROC) curve~\cite{Hoo:17} and the Precision–Recall (PR) curve~\cite{Cook:20}. The Area Under the ROC Curve (AUC) quantifies the trade-off between true positive rate and false positive rate, while the Average Precision (AP) summarizes precision across all recall levels.

Localization performance was evaluated using established object detection metrics~\cite{Rezatofighi2019_GIoU}. These include the Intersection over Union (IoU), the mean Average Precision (mAP), mAP at fixed IoU thresholds of 0.5 and 0.75 (mAP@0.5, mAP@0.75), and the mean Average Recall (mAR) at one and ten predictions per image (mAR@1, mAR@10). Since only a single defect class was considered, mean Average Precision reduces to Average Precision (mAP = AP).  

\section{Results}\label{sec:Results}
    The following section presents the evaluation of the zero-shot annotation pipeline using YOLOv8 models trained on bounding-box annotations. Quantitative performance metrics focus exclusively on these bounding-box-based models. In parallel, segmentation masks generated by SAM are shown for qualitative assessment, illustrating the propagation of errors from bounding boxes and the resulting mask quality. 
    
    % \begin{figure*}[t]
    %     \centering
    %     \includegraphics[width=\linewidth]{Figures/OPEX_yolo_models_pr_tpfp_curves_SHORT.pdf}
    %     \caption{ROC and PR curves for all trained YOLOv8 models using bounding boxes. Since the models achieve very high overall performance, only a selected portion of the curves is displayed to clearly illustrate the differences between the models.}%Given the overall high performance of the models, the axes have been adjusted and only a restricted portion of the curves is shown to enhance the visibility of relative differences.
    %     \label{fig:POCandPR}
    % \end{figure*}
  
    \begin{table*}[t]
    \caption{Classification and localization performance comparison of YOLOv8.}
    \label{tab:Loc}
    \centering
    %\resizebox{\textwidth}
    %\setlength{\tabcolsep}{} 
    \renewcommand{\arraystretch}{1.2} % etwas mehr Zeilenhöhe
    \begin{tabular}{l|cc|cccccc}
    \hline
     &  \multicolumn{2}{c|}{Classification} & \multicolumn{6}{c}{Localization}\\
    \textbf{Model} & \textbf{AUC} & \textbf{AP} & \textbf{IoU} & \textbf{mAP} & \textbf{mAP@50} & \textbf{mAP@75} & \textbf{mAR@1} & \textbf{mAR@10}\\
    \hline
    YOLOHand                    & 1.00 & 1.00 & 0.87 & 0.76 & 0.99 & 0.95 & 0.59 & 0.79 \\
    YOLOTwoCirclesBBoxDINO      & 0.98 & 0.98 & 0.67 & 0.30 & 0.62 & 0.26 & 0.29 & 0.44 \\
    YOLOTwoCirclesBBoxSAM       & 0.98 & 0.98 & 0.65 & 0.27 & 0.63 & 0.19 & 0.27 & 0.38 \\
    YOLOCellsBBoxDINO           & 0.98 & 0.98 & 0.60 & 0.18 & 0.38 & 0.15 & 0.24 & 0.36 \\
    YOLOCellsBBoxSAM            & 0.97 & 0.97 & 0.60 & 0.17 & 0.38 & 0.13 & 0.22 & 0.34 \\
    \hline
    \end{tabular}
    \end{table*}

    Table~\ref{tab:Loc} reports the binary classification and localization results of the YOLOv8 models trained with bounding-box annotations. 
    
    The reference model trained on manually annotated ground truth (YOLOHand) achieved the highest performance, with AUC and AP values close to one. Among the automatically generated label sets, the YOLOTwoCirclesBBoxDINO variant yielded the best results (AUC = 0.98, AP = 0.98). All models substantially exceeded the random baseline of 0.5, confirming reliable classification performance across prompt and annotation variants.

    \begin{figure*}[t]
        \centering
        \includegraphics[width=1.0\linewidth]{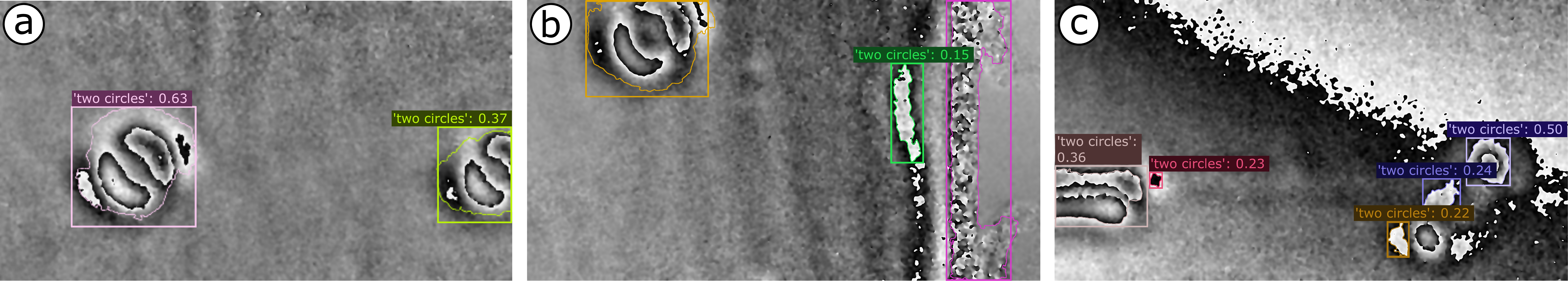}
        \caption{Bounding boxes and confidence scores predicted by Grounded DINO using the “Two Circles” prompt. (a) Example of correctly filtered detections. (b) Excessive bounding boxes around a streak-like artifact. (c) Right defect split into multiple boxes.}
        \label{fig:BBox-Seg-doubled}
    \end{figure*}
    
    The reference model trained on manually annotated ground truth (YOLOHand) achieved the highest scores across all metrics, with an IoU of 0.87 and an overall mAP of 0.76.
    Among the automatically generated datasets, models based on the Two Circles prompt obtained consistently higher scores than those derived from the Cells prompt. The highest values were observed for YOLOTwoCirclesBBoxDINO, which reached an IoU of 0.673 and an mAP of 0.300. The corresponding SAM-based bounding boxes (YOLOTwoCirclesBBoxSAM) achieved slightly reduced values, with IoU = 0.65 and mAP = 0.27. Models trained on the Cells prompt yielded the lowest scores overall, with IoU values around 0.60 and mAP values below 0.20, independent of whether DINO- or SAM-based annotations were used.
    
    % \begin{figure*}[t]
    %     \centering
    %     \includegraphics[width=1.0\linewidth]{Figures/OPEX_examples_smaller.pdf}
    %     \caption{Bounding boxes and confidence scores predicted by Grounded DINO using the “Two Circles” prompt. (a) Example of correctly filtered detections. (b) Excessive bounding boxes around a streak-like artifact. (c) Right defect split into multiple boxes.}
    %     \label{fig:BBox-Seg-doubled}
    % \end{figure*}
    
    \begin{figure*}[t]
        \centering
        \includegraphics[width=1.0\linewidth]{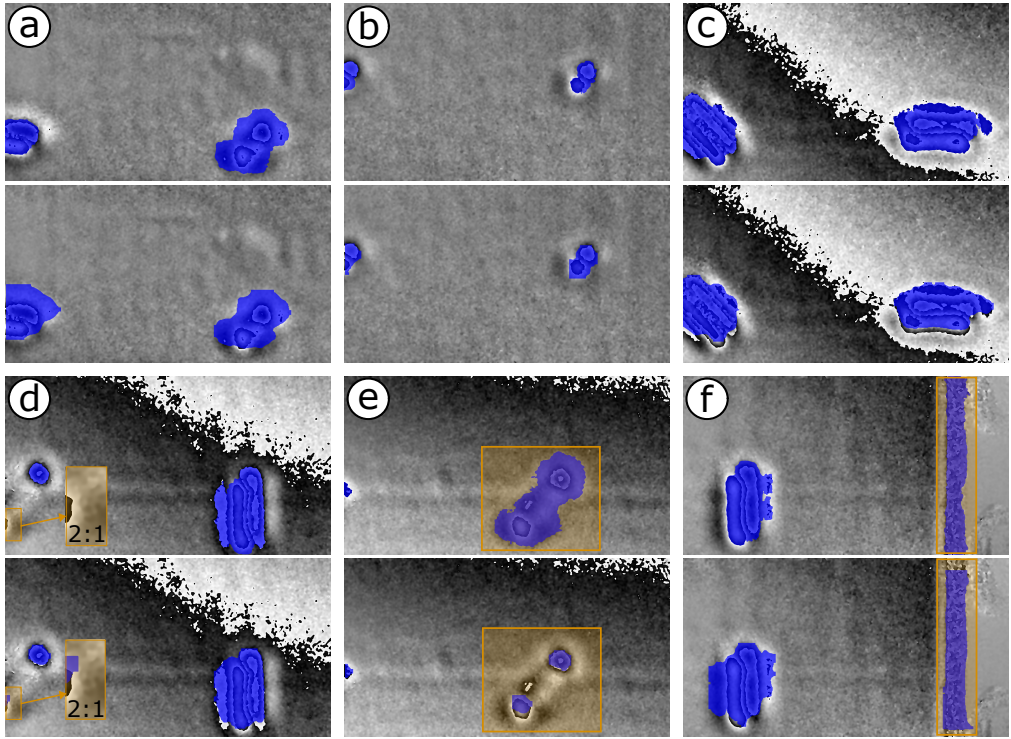}
        \caption{Validation results of the YOLO segmentation model trained on masks generated by the zero-shot Grounded-DINO + SAM pipeline using the “Two Circles“ prompt. Segmentation masks are blue, details are highlighted with orange boxes. For each case (a–d), the upper image shows the automatically generated segmentation mask, and the lower image shows the corresponding YOLO prediction. (a) illustrate the close correspondence between automatic labels and model output. (b) shows a defect region missed by the zero-shot pipeline but successfully detected by YOLO (detail scale 2:1); (c) shows a case where YOLO predicts a smaller, fragmented mask that no longer captures the full defect; and (d) demonstrates how an incorrectly placed automatic label is propagated through the trained model.}
        \label{fig:BBox-Seg-validation}
    \end{figure*}

    In addition to the quantitative evaluation, a qualitative inspection of the automatically generated labels was performed. Figure~\ref{fig:BBox-Seg-doubled} illustrates typical outcomes of the zero-shot pipeline using the Two Circles prompt. While correctly filtered bounding boxes and segmentation masks can be obtained (Fig.~\ref{fig:BBox-Seg-doubled}a), several sources of error were observed. These include excessive detections around streak-like artifacts (Fig.~\ref{fig:BBox-Seg-doubled}b) and duplicate bounding boxes describing the same defect (Fig.~\ref{fig:BBox-Seg-doubled}c). The impact of such errors on downstream training is exemplified in Fig.~\ref{fig:BBox-Seg-validation}, which compares the validation labels generated by DINO and SAM with the predictions of a YOLO model trained on this dataset. % Panel (a) shows a close correspondence between automatically generated labels and model predictions, while panels (b–d) illustrate special cases: defects missed by the automatic pipeline but detected by YOLO (b), fragmented predictions (c), and errors propagated from incorrectly generated labels (d). 
    
\section{Discussion}\label{sec:Discussion}
    The performance of the YOLO models trained on zero-shot annotations highlights both the potential and the limitations of the proposed pipeline. In the binary classification task, the Two Circles prompt combined with DINO-derived bounding boxes yielded the strongest results, underscoring the significant influence of prompt selection on downstream performance. This observation is particularly relevant in the context of shearography, where defect patterns are inherently difficult to describe even for human observers and thus lack clear semantic counterparts. 
    % Consequently, the selection and formulation of textual prompts remain a central design consideration: the choice of prompt can substantially influence the quality of automatically generated labels. At the same time, 
    This motivates complementary approaches, such as unsupervised anomaly-detection frameworks that operate on comparisons between defective and non-defective samples, which reduce reliance on human-defined semantic descriptors~\cite{Plassmann2025_SGAI}.
    
    For defect localization, the results demonstrate both the promise and the current limitations of zero-shot annotation. Despite the challenging nature of shearography images, YOLO models trained on automatically generated labels achieve consistent localization performance across different prompts. % The observed reduction in quantitative metrics such as IoU, mAP, and mAR should be interpreted with caution, as they are strongly influenced by characteristic error modes of Grounded DINO. Specifically, DINO tends to generate bounding boxes that are either overly small or fragmented into multiple parts. While these boxes often still capture the correct defect region, evaluation metrics only allow one bounding box to be matched to the ground truth, and additional boxes are penalized as false positives. This results in disproportionately lower scores, even when the defect itself is reliably detected. 
    
    Qualitative inspection of SAM segmentation masks confirms that the underlying spatial information is preserved, while metric degradation arises primarily from duplicated or oversegmented bounding boxes (Fig.~\ref{fig:BBox-Seg-doubled}).
    
    The integration of SAM adds a qualitatively richer annotation modality (Fig.~\ref{fig:BBox-Seg-doubled}), but the segmentation masks inherit the same structural biases from the underlying bounding boxes. This is further illustrated in Fig.~\ref{fig:BBox-Seg-validation}, where the trained YOLO model generally reproduces the SAM masks with high consistency (panel a), yet certain systematic effects, such as undetected defects, fragmented regions, or inaccurately placed annotations, remain observable (panels b-d). Consequently, while SAM enhances the visual interpretability of the annotations, it does not fully mitigate the metric degradation introduced by bounding-box duplication or oversegmentation.
    
    % Importantly, the performance differences between the Two Circles and Cells prompts (Table~\ref{tab:Loc}) further confirm that prompt design has a measurable influence on downstream localization. Nevertheless, both variants yield usable training data that enable YOLO to learn defect localization at a level that is remarkable given the absence of any manual annotation. 
    
    These results are further constrained by the fact that the zero-shot models employed were not specifically trained on shearography data, which reflects a general limitation of zero-shot methods in industrial applications~\cite{Lema2024_ZeroShotSegmentation}. % In addition, several heuristic post-processing steps were applied to filter and refine the generated annotations. The choice and configuration of these heuristics may influence the resulting labels, and systematic variation of these parameters could provide further insight into their impact on downstream model performance.
    
    Overall, the zero-shot annotation pipeline demonstrates that automated labeling can produce defect localization data of sufficient quality to train supervised models, even in a challenging domain such as shearography. While quantitative metrics may underestimate performance due to characteristic error modes, the approach provides a reproducible and scalable method to generate training data, reducing the need for extensive manual annotation.

\section{Conclusions and Outlook}\label{sec:Conclusions}
    This work presents a zero-shot annotation pipeline for shearography measurements, combining Grounded DINO and SAM to automatically generate defect labels. The pipeline enables the creation of training data for supervised defect detection models without manual intervention, addressing a major bottleneck in industrial NDT applications.
    
    Experimental evaluation demonstrates that the pipeline provides reliable binary classification and defect localization, with performance strongly influenced by prompt design and characteristic behaviors of the zero-shot detector. 
    %The Two Circles prompt consistently outperformed the Cells prompt, highlighting the importance of semantic specificity in text-based label generation. Despite metric limitations introduced by bounding-box fragmentation and duplication, qualitative inspection confirms that defect regions are accurately captured.
    These findings indicate that zero-shot annotation can complement or partially replace manual labeling in industrial shearography, providing reproducible, scalable training data. 
    
    With the recent SAM 3 release, exemplar-based prompting has become available as an additional conditioning mechanism. Since this modality differs fundamentally from the box-based setup underlying the present pipeline, it was not included in this study. However, evaluating whether exemplar prompts can capture defect-specific structures more effectively represents a promising direction for future research.
    
    % Future work could also investigate domain-adaptive fine-tuning of zero-shot models and systematic optimization of heuristic post-processing. Pipelines incorporating preliminary anomaly detection to identify clusters of defects could iteratively refine textual prompts, further improving zero-shot label quality. Extending the approach to multi-class settings would enable differentiation and categorization of distinct defect types, supporting more detailed defect characterization in industrial applications.

\begin{acknowledgement}
The authors would like to thank the Ministry of Science and Health Rhineland-Palatinate and Trier University of Applied Sciences for supporting the publication of this research through the Young Researchers Fund. We also acknowledge the BMWK's ZIM program, which made the employment of researchers such as Ms. Plassmann possible. Furthermore, we extend our appreciation to Tenta Vision GmbH for providing additional equipment, which complemented our own resources and helped strengthen the validity of our simulation results. We also thank our colleagues for their valuable feedback and support during this research.
\end{acknowledgement}
\bibliographystyle{spmpsci}  % Alternativen: spbasic, spphys
\bibliography{literature.bib}

\end{document}